\documentclass[runningheads]{llncs}

\usepackage{eccv}

\usepackage{eccvabbrv}

\usepackage{graphicx}
\usepackage{booktabs}

\usepackage[accsupp]{axessibility}  % Improves PDF readability for those with disabilities.

\usepackage{hyperref}

\usepackage{orcidlink}

\usepackage{url}

\usepackage{algorithm}
\usepackage{algorithmic}

\usepackage{multirow}
\usepackage{adjustbox}
\usepackage{makecell}
\usepackage{amsmath}
\usepackage{amssymb}
\usepackage{bm}
\usepackage{xcolor}
\usepackage{marvosym}

\makeatletter
\def\blfootnote{\gdef\@thefnmark{}\@footnotetext}
\makeatother

\begin{document}

\title{ReynoldsFlow: Physics-Inspired Spatiotemporal Flow Representation for Video Understanding}

\titlerunning{ReynoldsFlow}

\author{Yu-Hsi Chen\inst{1}\orcidlink{0009-0006-1771-0289} \and
Ching-Kai Lin\inst{2}\orcidlink{0009-0005-5226-6032} \and
PingKong Huang\inst{2}\orcidlink{0009-0007-1358-6468} \and
Chin-Tien Wu\inst{2,}\textsuperscript{\Letter}\orcidlink{0000-0001-8755-840X}
}

\authorrunning{Y.-H. Chen et al.}

\institute{The University of Melbourne, Parkville, Australia \and
Department of Applied Mathematics, National Yang Ming Chiao Tung University, Hsinchu 300, Taiwan
\email{ctw@math.nctu.edu.tw}}

\maketitle
\blfootnote{\Letter~Corresponding author.}

% ================ Abstract ================ %
\begin{abstract}
Video understanding has largely relied on deep spatiotemporal architectures, including 3D convolutional networks and optical flow (OF) based models. While effective, these methods are often computationally expensive and depend on heuristic motion representations that are sensitive to illumination, scale, and structural changes. To address these limitations, we propose ReynoldsFlow, a physics-inspired representation grounded in the Reynolds transport theorem (RTT) and Helmholtz-Hodge decomposition (HHD). ReynoldsFlow decomposes motion into curl-free (CF) and divergence-free (DF) components, providing a principled and interpretable characterization of scene dynamics. By coupling intensity information with decomposed motion cues, it produces dynamics-aware, texture-preserving features that boost downstream tasks such as pose estimation, action recognition, and tiny object detection. Lightweight and modular, ReynoldsFlow can be readily integrated into existing architectures. Experiments across diverse benchmarks show that ReynoldsFlow consistently matches or surpasses existing approaches, offering improved generalizability and computational efficiency.
\end{abstract}

% ================ Introduction ================ %
\section{Introduction}
\label{sec:introduction}
Computer vision is increasingly integrated into daily life, with video functionality central to modern smartphones. Beyond casual use, video understanding supports a wide range of applications, including stabilization, interpolation, object detection~\cite{jiao2021new,zhu2018towards,zhu2020review,zhu2017flow}, multi-object tracking~\cite{ciaparrone2020deep,agbinya1999multi,chen2025strong}, and pose estimation~\cite{girdhar2018detect,charles2016personalizing,von2016human,pavllo20193d}. Progress has been largely driven by deep learning (DL) architectures such as RNNs~\cite{zhao2017hierarchical,ebrahimi2015recurrent}, LSTMs~\cite{graves2012long,zhang2016video}, and, more recently, transformer- and Mamba-based spatiotemporal models~\cite{tran2024transformer,hsu2023video,gai2023spatiotemporal,li2024stnmamba,park2024videomamba}. Despite their strong empirical performance, these methods remain computationally expensive, rely on carefully tuned heuristics, and offer limited interpretability, as their representations are rarely grounded in physical motion principles.

To address these challenges, we propose ReynoldsFlow, a physics-inspired video representation. Combining OF~\cite{sun2010secrets}, RTT~\cite{white2011fluid}, and the HHD~\cite{arfken2011mathematical}, ReynoldsFlow produces interpretable spatiotemporal features in a training-free, unsupervised manner. By deriving video representations directly from data rather than from heuristic motion representations, the resulting features are more efficient and generalizable. Our main contributions are summarized as follows:
\begin{itemize}
    \item[1.] \textbf{ReynoldsFlow formulation.} We propose ReynoldsFlow, an unsupervised motion representation that decomposes motion into CF and DF components via RTT and HHD, providing physically interpretable dynamics.
    \item[2.] \textbf{ReynoldsFlow representation.} We provide a visualization that integrates decomposed motion magnitudes with image appearance cues, yielding a more interpretable and robust representation than HSV-based visualization.
    \item[3.] \textbf{Comprehensive evaluation.} We validate ReynoldsFlow on multiple video benchmarks, showing its effectiveness for video representation learning.
\end{itemize}

% ================ Related Work ================ %
\section{Related Work}
\label{sec:related_work}

% ======== Optical Flow and Video Motion Estimation ======== %
\subsection{Optical Flow and Video Motion Estimation}
\label{ssec:optical_flow_and_video_motion_estimation}

Early OF methods established the foundational tradeoff between local accuracy and global smoothness, pioneered by the works of Lucas-Kanade (LK)~\cite{lucas1981iterative} and Horn-Schunck (HS)~\cite{horn1981determining}. Subsequent classical approaches focused on overcoming the fragility of these formulations. Variational methods and polynomial expansion algorithms~\cite{farneback2003two, brox2004high} significantly improved sub-pixel precision, while others prioritized robustness against illumination changes and noise~\cite{zach2007duality, tao2012simpleflow}. To handle large displacements and complex occlusions, later heuristic techniques integrated sparse feature matching with dense interpolation~\cite{weinzaepfel2013deepflow, wulff2015efficient, kroeger2016fast}. While interpretable and mathematically grounded, these techniques can be less effective when handling complex real-world motion under computational constraints.

The paradigm subsequently shifted toward data-driven architectures. CNN-based frameworks demonstrated the viability of end-to-end prediction~\cite{dosovitskiy2015flownet}, though they initially lacked fine-grained detail. This limitation drove a wave of architectural innovations, including cascaded refinement~\cite{ilg2017flownet}, cost-volume pyramids~\cite{sun2018pwc, ranjan2017optical}, and edge-aware sparse matching~\cite{revaud2015epicflow}. This evolution culminated in models such as RAFT~\cite{teed2020raft}, which set a new standard by iteratively refining dense flows using learned contextual correlations. Recent extensions have further pushed the boundaries of temporal consistency and cross-domain generalization~\cite{morimitsu2024recurrent, wang2024sea, morimitsu2025dpflow}. However, despite their strong performance, deep OF networks typically require large-scale training data and task-specific optimization.

Application-driven integrations of OF highlight these practical limitations. For UAV detection,~\cite{sun2023enhancing} embeds an LK-inspired layer into YOLO, while~\cite{madake2023golf} leverages Farneback OF for golf pose estimation using handcrafted features. Both demonstrate ingenuity but remain constrained by the limited expressiveness of flow visualizations. To our knowledge, no existing work effectively leverages mathematically rigorous, physics-aware OF images as direct network inputs to substantially improve downstream detection or action recognition tasks.

% ======== Decomposition-based Representations ======== %
\subsection{Decomposition-based Representations}
\label{ssec:decomposition-based_representations}

The HHD provides a principled mathematical framework for separating a velocity field into CF, DF, and harmonic components~\cite{schwarz2006hodge,bhatia2012helmholtz}. This formulation has been widely applied in computer vision and graphics for interpreting flow fields and solving boundary value problems. Early work demonstrated applications of discrete HHD to image processing, enabling structural analysis of visual data~\cite{palit2005applications}. To enhance robustness, convex optimization formulations were later introduced to regularize image flows and stabilize the decomposition process~\cite{yuan2009convex}. Beyond motion analysis, decomposition-based formulations have found broad applications in image representation, enabling structural organization of large-scale image databases~\cite{guo1997image}, signal separation through sparse decomposition~\cite{fadili2009image}, and multimodal image fusion via decomposition and sparse coding~\cite{zhu2018novel}.

More recently, decomposition-based techniques have been extended to practical vision tasks, such as illumination compensation and texture enhancement in imaging~\cite{wu2020illuminance}. Physics-inspired neural architectures further push this line of research, as in HDNet~\cite{qi2024hdnet}, which leverages HHD to design interpretable deep networks for flow estimation. Collectively, these works show that decomposition-based representations can improve interpretability and robustness in visual analysis. However, most approaches remain confined to spatial formulations, focusing on single-frame decomposition of images or velocity fields while neglecting temporal dynamics. This lack of spatiotemporal integration limits their applicability to modern video understanding tasks, where temporal coherence is critical.

% ======== Spatiotemporal Modules in Video Learning ======== %
\subsection{Spatiotemporal Modules in Video Learning}
\label{ssec:spatiotemporal_modules_in_video_learning}

Modeling temporal dynamics is fundamental to video understanding, as motion provides critical cues for object behavior, scene evolution, and activity recognition. To capture such dynamics, DL methods incorporated 3D convolutions and recurrent architectures to learn spatiotemporal representations directly from video sequences~\cite{xie2017rethinking,qiu2017learning,peng2021deep,ul2024video}. However, their reliance on large-scale supervised training and computationally intensive optimization often limits efficiency and generalization in data-scarce settings.
Physics-inspired approaches offer a principled alternative to purely data-driven temporal modeling. Previous work employed the RTT to treat video sequences as continuous frame flows, enabling theoretically grounded analysis of temporal changes~\cite{shih2001shot}. Building on this idea, ReynoldsFlow combines RTT with the HHD to decompose motion into two flow components as an unsupervised approach. This method bridges principled physical modeling with modern feature learning, producing generalizable representations without task-specific training.

% ================ ReynoldsFlow ================ %
\section{ReynoldsFlow}
\label{sec:reynoldsflow}
In this section, we formulate ReynoldsFlow based on RTT and HHD, establishing a physics-inspired framework for velocity field estimation. We further present the visualization scheme and analyze its computational efficiency.

% ======== Preliminaries ======== %
\subsection{Preliminaries}

% ==== Helmholtz-Hodge decomposition ==== %
\subsubsection{Helmholtz-Hodge decomposition}
Let \(\bm{v} \in C^1(\Omega, \mathbb{R}^2)\) be a continuously differentiable velocity field defined on a domain $\Omega$. According to the HHD theorem~\cite{arfken2011mathematical}, \(\bm{v}\) can be uniquely decomposed into the sum of a CF (irrotational) component \(\bm{v}_c\) and a DF (solenoidal) component \(\bm{v}_d\):
\begin{equation}
    \label{eq:helmholtz_decomposition}
    \bm{v} = \bm{v}_c + \bm{v}_d,
\end{equation}
where \(\bm{v}_c\) satisfies \( \nabla \times \bm{v}_c = 0 \) and \(\bm{v}_d\) satisfies \( \nabla \cdot \bm{v}_d = 0 \).

% ==== Reynolds transport theorem ==== %
\subsubsection{Reynolds transport theorem}
Consider a smooth scalar function \(f=f(\bm{p},t)\) defined on a time-dependent domain \(\Omega(t)\). A grayscale video is represented as \(f(\bm{p}(t^n),t^n)\), where \(f\) denotes the intensity at pixel location \(\bm{p}(t^n)\) in the domain \(\Omega(t^n)\), and \(t^n\) is the timestamp of the \(n\)-th frame. The RTT~\cite{white2011fluid} states that:
\begin{equation}
    % \resizebox{\linewidth}{!}{$
    \begin{aligned}
    \frac{d}{dt}\int_{\Omega(t)} f \, dA &= \int_{\Omega(t)} \frac{\partial f}{\partial t} \, dA + \int_{\partial \Omega(t)} f (\bm{v} \cdot \bm{n}) \, dS \\
    &= \int_{\Omega(t)} \frac{\partial f}{\partial t} \, dA + \int_{\Omega(t)} \nabla \cdot (f \bm{v}) \, dA \\
    &= \int_{\Omega(t)} \left( \frac{\partial f}{\partial t} + \nabla f \cdot \bm{v} + f \nabla \cdot \bm{v} \right) \, dA.
    \end{aligned}
    \label{eq:RT}
    % $}
\end{equation}
where \(\bm{v} = \bm{v}(\bm{p}, t)\) is the velocity field at $\bm{p} \in \Omega(t)$ and time \(t\). For video footage, if the brightness in the region $\int_{\Omega(t)} f dA$ is constant and the velocity field \(\bm{v}\) is DF (i.e., $\bm{v}=\bm{v}_d$) for all time, then~\cref{eq:RT} reduces to:
\begin{equation}
    0 = \int_{\Omega(t)} \left( \frac{\partial f}{\partial t} + \nabla f \cdot \bm{v}_d \right) \, dA.
    \label{eq:Opt_flow}
\end{equation}
Assuming the integrand in~\cref{eq:Opt_flow} vanishes pointwise, and $\bm{v}_d$ is piecewise constant,~\cref{eq:Opt_flow} reduces to the traditional OF.

% ==== Area Jacobian in domain transformation ==== %
\subsubsection{Area Jacobian in domain transformation}
Consider a region \(\Omega(t)\) evolving over time \(t\) under the influence of a velocity field \(\bm{v}\). Let \(\bm{p}^n = \begin{pmatrix} x^n \\ y^n \end{pmatrix} \in \Omega^n = \Omega(t^n)\) at time \(t^n\), with corresponding velocity field \(\bm{v}^n = \begin{pmatrix} v_x^n \\ v_y^n \end{pmatrix}\), and assume a uniform time increment \(\Delta t = t^{n+1} - t^n\). Using the explicit Euler method, the domain transformation from \(\Omega^n\) to \(\Omega^{n+1}\) can be approximated by:
\begin{equation}
    \label{eq:position}
    \bm{p}^{n+1} \approx \bm{p}^n + \bm{v}^n(\bm{p}^n) \Delta t.
\end{equation}
From~\cref{eq:position}, the differential wedge product \(dx \wedge dy\) at \(\bm{p}^{n+1}\) can be written as:
\begin{equation*}
    \begin{aligned}
    dx^{n+1} \wedge dy^{n+1} &\approx \left(dx^n + dv_x^n \Delta t\right) \wedge \left(dy^n + dv_y^n \Delta t\right)\\
    &= dx^n \wedge dy^n + dx^n \wedge dv_y^n \Delta t + dv_x^n \Delta t \wedge dy^n + dv_x^n \Delta t \wedge dv_y^n \Delta t.
    \end{aligned}
\end{equation*}
Using the fact that
\begin{equation*}
    dv_x = \frac{\partial v_x}{\partial x} dx + \frac{\partial v_x}{\partial y} dy, \quad
    dv_y = \frac{\partial v_y}{\partial x} dx + \frac{\partial v_y}{\partial y} dy,
\end{equation*}
we obtain
\begin{equation*}
    dx^{n+1} \wedge dy^{n+1} = dx^n \wedge dy^n + \nabla \cdot \bm{v}^n \, dx^n \wedge dy^n \, \Delta t + \mathcal{O}((\Delta t)^2).
\end{equation*}
For sufficiently small \(\Delta t\), this simplifies to:
\begin{equation}
    \label{eq:area_transformation}
    dx^{n+1} \wedge dy^{n+1} \approx \left(1 + \nabla \cdot \bm{v}^n \Delta t \right) dx^n \wedge dy^n.
\end{equation}
As a result, \(\left(1 + \nabla \cdot \bm{v}^n \Delta t \right)\) approximates the Jacobian determinant relating area elements \(dA^n\) in \(\Omega^n\) to \(dA^{n+1}\) in \(\Omega^{n+1}\). In other words,
\begin{equation}
    \label{eq:Jacobian}
    dA^{n+1} \approx \left(1 + \nabla \cdot \bm{v}^n \Delta t \right) dA^n.
\end{equation}

% ======== CF Component: Irrotational Field ======== %
\subsection{CF Component: Irrotational Field}
\label{ssec:irrotational_component}
To generalize beyond the brightness constancy and DF assumptions, we invoke the RTT on each local patch $\omega(t)$ combined with the HHD~\cref{eq:helmholtz_decomposition} to rewrite:
\begin{equation}
\label{eq:RT_H_decomp}
    \frac{d}{dt}\int_{\omega(t)} f \, dA 
    = \int_{\omega(t)} \left( \frac{\partial f}{\partial t} + \nabla f \cdot \bm{v}_d + \nabla f \cdot \bm{v}_c + f \nabla \cdot \bm{v}_c \right) \, dA.
\end{equation}
Applying the explicit Euler method, the LHS of~\cref{eq:RT_H_decomp} is approximated as:
\begin{equation}
\label{eq:lhs}
   \begin{aligned}
 \text{LHS} &= \frac{1}{\Delta t} \left( \int_{\omega^{n+1}}f^{n+1} \, dA^{n+1} - \int_{\omega^n}f^n \, dA^n \right), \quad \\
         &\approx \frac{1}{\Delta t} \int_{\omega^n} \left[ (1 + \nabla \cdot \bm{v}^n \Delta t) f^{n+1} - f^n \right] \, dA^n, \quad \text{by~\cref{eq:Jacobian}}\\
         &= \int_{\omega^n} \left(\frac{f^{n+1} - f^n}{\Delta t} + f^{n+1}\nabla \cdot \bm{v}^n \right) \, dA^n.
   \end{aligned}
\end{equation}
Next, using the Taylor approximation:
\[
 f^{n+1}-f^n \approx \frac{\partial f^n}{\partial t}\Delta t + \nabla f^n \cdot \left(\bm{v}^n \Delta t \right) =  \frac{\partial f^n}{\partial t}\Delta t + \nabla f^n \cdot \begin{pmatrix}
   \Delta x^n \\ \Delta y^n
   \end{pmatrix},
\] and the HHD of the velocity field $\bm{v}$,
\begin{equation*}
      \begin{aligned}
 \text{LHS} &\approx \int_{\omega^n} \left(\frac{\partial f^n}{\partial t} + \nabla f^n \cdot \bm{v}^n + f^{n+1}\nabla \cdot \bm{v}^n \right)\, dA^n,\\
      &= \int_{\omega^n} \left(\frac{\partial f^n}{\partial t} + \nabla f^n \cdot (\bm{v}_c^n + \bm{v}_d^n) + f^{n+1}\nabla \cdot \bm{v}_c^n \right) dA^n.
      \end{aligned}
\end{equation*}
Similarly, the RHS of~\cref{eq:RT_H_decomp} is
\begin{equation}
\label{eq:rhs}
    \text{RHS}= \int_{\omega^n} \left(\frac{\partial f^n}{\partial t} + \nabla f^n \cdot (\bm{v}_c^n +  \bm{v}_d^n) + f^{n}\nabla \cdot \bm{v}_c^n\right)\, dA^n,
\end{equation}
Equating both sides and defining \(\delta f^n = f^{n+1} - f^n\) gives
\begin{equation*}
    \int_{\omega^n} \delta f^n \nabla \cdot \bm{v}_c^n \, dA^n = 0.
\end{equation*}
Integration by parts leads to the variational form:
\begin{equation}
   \label{eq:goal}
   \int_{\partial \omega^n} \delta f^n \bm{v}_c^n \cdot \bm{n} \, dS^n - \int_{\omega^n} \nabla \delta f^n \cdot \bm{v}_c^n \, dA^n = 0.
\end{equation}

We can compute the velocity field $\bm{v}_c^n$ from~\cref{eq:goal} by assuming it remains constant within each local window patch $\omega^n$. Specifically, we compute $\bm{v}_c^n$ on a \(3 \times 3\) window patch, denoted as \(\omega^n_{3\times 3}\). To approximate the boundary integral term in~\cref{eq:goal}, we apply Simpson's rule:
\begin{equation}
   \int_{\partial \omega_{3\times 3}^n} \delta f^n \bm{v}_c^n \cdot \bm{n} \, dS^n \approx \left[ (\delta f^n_{b})_x, (\delta f^n_{b})_y \right] \cdot \bm{v}_c^n,
   \label{eq:bdy_int}
\end{equation}
where 
\begin{equation*}
    (\delta f^n_{b})_x = \frac{1}{3} \begin{bmatrix}
    -1 &  0 &  1\\
    -4 &  0 &  4\\
    -1 &  0 &  1
    \end{bmatrix}
    \ast \delta f^n  \: \text{and} \:
    (\delta f^n_{b})_y = \frac{1}{3} \begin{bmatrix}
    1 &  4 &  1\\
    0 &  0 &  0\\
    -1 & -4 & -1
    \end{bmatrix}
    \ast \delta f^n.
\end{equation*} 
The domain integral term in~\cref{eq:goal} is approximated as:
\begin{equation}
    \int_{\omega^n_{3\times 3}} \nabla \delta f^n \cdot \bm{v}_c^n \, dA^n = \int_{\omega^n} [(\nabla \delta f^n)_x, (\nabla \delta f^n)_y] \cdot \bm{v}_c^n\, dA^n \approx [(\nabla \delta f^n_\omega)_x, (\nabla \delta f^n_\omega)_y] \cdot \bm{v}_c^n ,
    \label{eq:dom_int}
\end{equation}
where
\begin{equation*}
    \resizebox{\linewidth}{!}{$
    (\nabla \delta f^n_\omega)_x= 
    \begin{bmatrix}
    1 & 1 & 1\\
    1 & 1 & 1\\
    1 & 1 & 1
    \end{bmatrix}
    \ast \left(\begin{bmatrix}
    -1 & 0 & 1 \\
    -2 & 0 & 2 \\
    -1 & 0 & 1
    \end{bmatrix} \ast \delta f^n \right) \: \text{and} \:
    (\nabla \delta f^n_{\omega})_y=\begin{bmatrix}
    1 & 1 & 1\\
    1 & 1 & 1\\
    1 & 1 & 1
    \end{bmatrix} \ast \left(\begin{bmatrix}
    -1 & -2 & -1 \\
    0 &  0 &  0 \\
    1 &  2 &  1
    \end{bmatrix} \ast \delta f^n \right).
    $}
\end{equation*}
By substituting the discrete approximations~\cref{eq:bdy_int} and~\cref{eq:dom_int} into the variational form~\cref{eq:goal}, we derive the irrotational velocity field \(\bm{v}_c^n\). To ensure spatial smoothness and robustness against local noise, we apply a Gaussian smoothing kernel \(G\), defining the CF velocity field as:
\begin{equation}
    \label{eq:reynoldsflow}
    \bm{v}_c^n = \begin{bmatrix}
    G \ast \left( (\delta f^n_{b})_x - (\nabla \delta f^n_{\omega})_x \right)\\
    G \ast \left(  (\delta f^n_{b})_y - (\nabla \delta f^n_{\omega})_y \right)
    \end{bmatrix}.
\end{equation}
This irrotational velocity field \(\bm{v}_c^n\) isolates the residual expansion fluxes, non-rigid deformations, and varying illuminations. We can then plug \(\bm{v}_c^n\) back into the transport equation to resolve the complete motion field.

% ======== DF Component: Solenoidal Field ======== %
\subsection{DF Component: Solenoidal Field}
\label{ssec:solenoidal_component}
To resolve the DF component $\bm{v}_d^n$, we equate the discrete formulation in~\cref{eq:lhs} with the continuous formulation in~\cref{eq:rhs}:
\begin{equation}
\label{eq:equate_lhs_rhs}
\resizebox{\linewidth}{!}{$
\int_{\omega^n} \left(\frac{f^{n+1} - f^n}{\Delta t} + f^{n+1}\nabla \cdot \bm{v}_c^n \right) dA^n
=
\int_{\omega^n} \left(\frac{\partial f^n}{\partial t} + \nabla f^n \cdot (\bm{v}_c^n + \bm{v}_d^n) + f^n\nabla \cdot \bm{v}_c^n \right) dA^n.
$}
\end{equation}
The discrete temporal difference can be related to the continuous derivative using the Taylor expansion
\[
f^{n+1} = f^n + \frac{\partial f^n}{\partial t}\Delta t
+ \frac{\partial^2 f^n}{\partial t^2}\frac{(\Delta t)^2}{2}
+ \mathcal{O}((\Delta t)^3).
\]
Ignoring higher-order terms gives
\begin{equation*}
\frac{f^{n+1}-f^n}{\Delta t} - \frac{\partial f^n}{\partial t}
\approx
\frac{\partial^2 f^n}{\partial t^2}\frac{\Delta t}{2} .
\end{equation*}
Substituting this relation into~\cref{eq:equate_lhs_rhs} and isolating the advection term yields
\begin{align}
\int_{\omega^n} \nabla f^n \cdot \bm{v}_d^n \, dA^n
&\approx
\int_{\omega^n}
\left(
(f^{n+1}-f^n)\nabla \cdot \bm{v}_c^n
-
\nabla f^n \cdot \bm{v}_c^n
+
\frac{\partial^2 f^n}{\partial t^2}\frac{\Delta t}{2}
\right)dA^n \nonumber \\
&\approx
\int_{\omega^n}
\left(
\delta f^n \nabla \cdot \bm{v}_c^n
-
\nabla f^n \cdot \bm{v}_c^n
+
\frac{f^{n+1}-2f^n+f^{n-1}}{2\Delta t}
\right)dA^n,
\label{eq:vd_integral_generalized}
\end{align}
where the second-order temporal derivative is approximated using central differencing.
Although~\cref{eq:vd_integral_generalized} captures intensity acceleration, the second-order term is noise-sensitive and vanishes under constant-velocity translation. We therefore impose the assumption of local intensity preservation, where the total intensity within the moving patch remains approximately constant:
\[
\int_{\omega^n} f^{n+1} dA^n \approx \int_{\omega^n} f^n dA^n .
\]
In the absence of geometric divergence, $\bm{v}_c^n=0$, the discrete LHS of~\cref{eq:equate_lhs_rhs} becomes zero. Therefore, physical consistency requires:
\begin{equation}
\int_{\omega^n}
\left(
\frac{\partial f^n}{\partial t}
+
\nabla f^n \cdot (\bm{v}_c^n + \bm{v}_d^n)
+
f^n \nabla \cdot \bm{v}_c^n
\right)dA^n
=0.
\end{equation}
By the localization theorem, the integrand vanishes pointwise. Approximating $\frac{\partial f^n}{\partial t} \approx \delta f^n$ removes the acceleration term while preserving first-order consistency. Rearranging gives the divergence-compensated residual map $D$:
\begin{equation}
\label{eq:residual_map}
\nabla f^n \cdot \bm{v}_d^n
=
\underbrace{-\delta f^n}_{\text{Temporal Diff.}}
-
\underbrace{\nabla f^n \cdot \bm{v}_c^n}_{\text{CF Advection}}
-
\underbrace{f^n \nabla \cdot \bm{v}_c^n}_{\text{CF Dilation}}
\equiv D.
\end{equation}
When $\bm{v}_c^n=0$,~\cref{eq:residual_map} naturally reduces to the classical brightness constancy constraint
$\nabla f^n \cdot \bm{v}_d^n = -\delta f^n$, which corresponds to~\cref{eq:Opt_flow}.

To estimate the DF velocity field $\bm{v}_d^n=(u_d,v_d)^\top$, we assume it is locally constant within a neighborhood $\omega$. Following the LK framework~\cite{lucas1981iterative,barron1994performance}, we solve a Gaussian-weighted least squares problem:
\begin{equation}
\min_{\bm{v}_d^n}
\sum_{\bm{x}\in\omega}
W(\bm{x})
\left(
\nabla f^n(\bm{x})\cdot \bm{v}_d^n - D(\bm{x})
\right)^2 ,
\end{equation}
where $W(\bm{x})$ is a Gaussian weighting function. This yields
\begin{equation}
\begin{bmatrix}
\sum W (f_x^n)^2 & \sum W f_x^n f_y^n \\
\sum W f_x^n f_y^n & \sum W (f_y^n)^2
\end{bmatrix}
\begin{bmatrix}
u_d \\ v_d
\end{bmatrix}
=
\begin{bmatrix}
\sum W f_x^n D \\
\sum W f_y^n D
\end{bmatrix},
\label{eq:reynolds_matrix}
\end{equation}
where $f_x^n$ and $f_y^n$ denote spatial gradients.
Finally, the complete ReynoldsFlow is formulated as \(\bm{v}^n_R = \bm{v}_c^n + \bm{v}_d^n\).

% ======== ReynoldsFlow Representation ======== %
\subsection{ReynoldsFlow Representation}
\label{ssec:visualization}
\begin{figure}[tb]
    \begin{center}
    \includegraphics[width=1\linewidth]{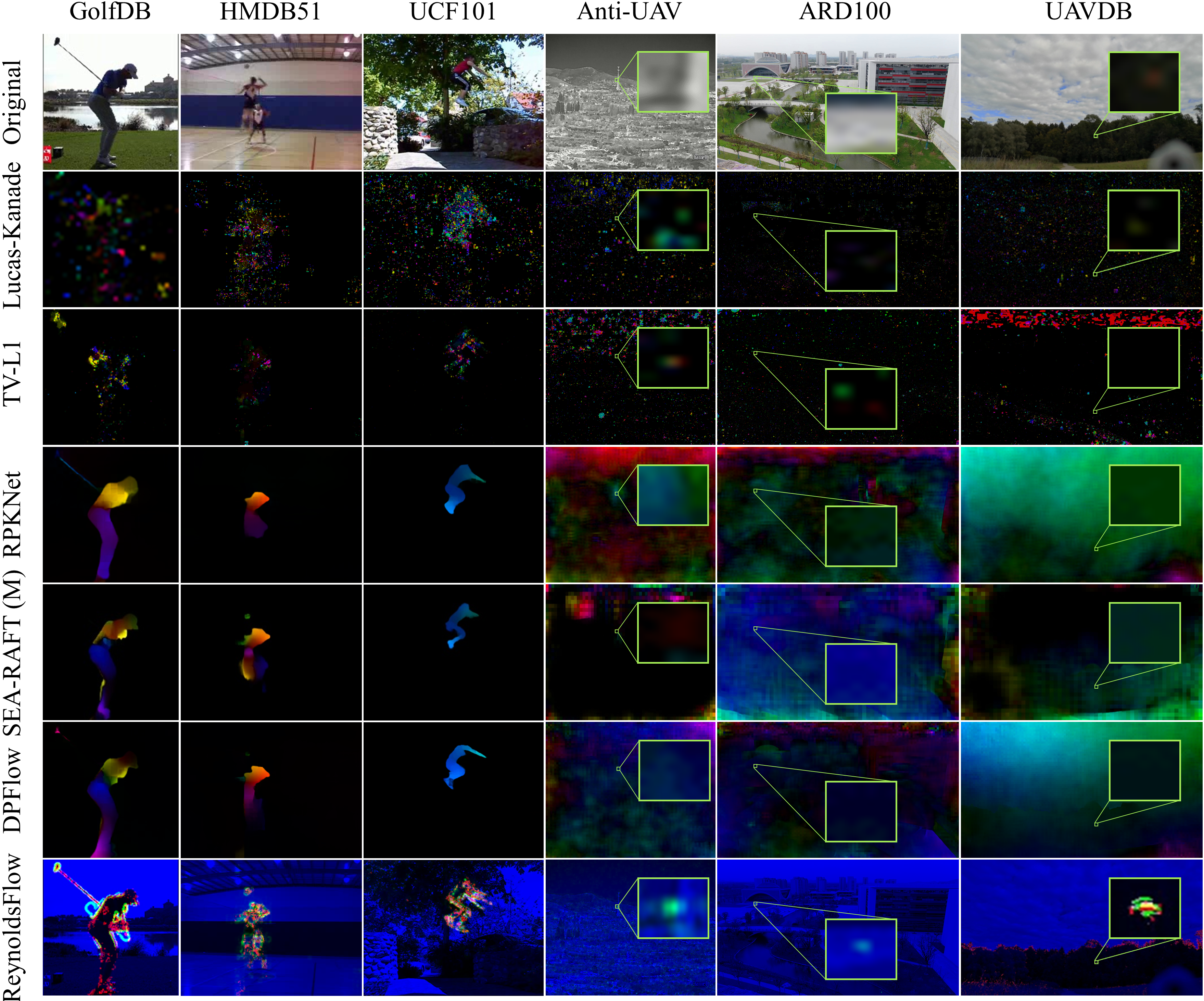}
    \end{center}
    \caption{Top row: Sample frames from six benchmark datasets. Middle rows: HSV visualizations from five OF methods. Bottom row: ReynoldsFlow representations.}
    \label{fig:flow_comparison}
\end{figure}

Flow is commonly visualized in the HSV color space, where motion magnitude and direction are encoded as hue and saturation~\cite{baker2011database}. However, the nonlinear HSV-to-RGB transformation can introduce perceptual inconsistencies, particularly in low-texture regions, complex illumination, or dynamic motion scenarios. These limitations may degrade the reliability of flow representations for downstream tasks such as tiny object detection.
To address this issue, we provide an alternative visualization that augments flow with additional appearance cues. Specifically, we construct a three-channel representation
\[
\bm{F}^n_{\text{R}} = \left[|\bm{v}_d^n|,\ |\bm{v}_c^n|,\ f^n\right],
\]
where the red and green channels encode the magnitudes of the DF and CF, respectively, while the blue channel preserves the current frame intensity $f^n$.
By integrating motion magnitudes with image appearance, this representation improves velocity field interpretability and robustness to visual ambiguity, particularly for tiny or fast-moving objects. As shown in~\cref{fig:flow_comparison}, $\bm{F}^n_{\text{R}}$ produces sharper and more stable features across diverse datasets.

% ================ Experimental Results ================ %
\section{Experimental Results}
\label{sec:results}
We first evaluate ReynoldsFlow against classical and DL-based OF methods under four controlled scenarios: (1) geometric divergence, (2) illumination variation, (3) horizontal translation, and (4) pure rotation.
We then assess the ReynoldsFlow representation on downstream vision tasks, including pose estimation on GolfDB~\cite{9025559}, action recognition on HMDB51~\cite{kuehne2011hmdb} and UCF101~\cite{soomro2012ucf101}, and object detection on Anti-UAV~\cite{jiang2021anti}, ARD100~\cite{guo2025yolomg}, and UAVDB~\cite{chen2024uavdb}.

We compare ReynoldsFlow with 13 input representations: (1) RGB, (2) grayscale, classical OF methods including (3) Horn-Schunck~\cite{horn1981determining}, (4) Lucas-Kanade~\cite{lucas1981iterative}, (5) Farneback~\cite{farneback2003two}, (6) Brox~\cite{brox2004high}, (7) TV-L1~\cite{zach2007duality}, (8) DeepFlow~\cite{weinzaepfel2013deepflow}, (9) PCAFlow~\cite{wulff2015efficient}, and (10) DIS~\cite{kroeger2016fast}, as well as DL-based methods (11) RPKNet~\cite{morimitsu2024recurrent}, (12) SEA-RAFT~\cite{wang2024sea}, and (13) DPFlow~\cite{morimitsu2025dpflow}. Classical methods were chosen for their training-free nature, consistent with ReynoldsFlow, while DL-based methods used official pretrained models due to the absence of ground-truth velocity fields in most datasets.
All velocity fields are visualized in HSV following~\cite{liu2009beyond}, where hue and saturation encode motion direction and magnitude, respectively. ReynoldsFlow follows the visualization scheme in~\cref{ssec:visualization}.

% ======== Controlled Evaluation of Decomposed Motion Fields ======== %
\subsection{Controlled Evaluation of Decomposed Motion Fields}
\label{ssec:controlled_scenarios}

% ==== Setup ==== %
\subsubsection{Setup}
\label{sssec:experimental_setup}
\begin{figure}[tb]
    \centering
    \includegraphics[width=\linewidth]{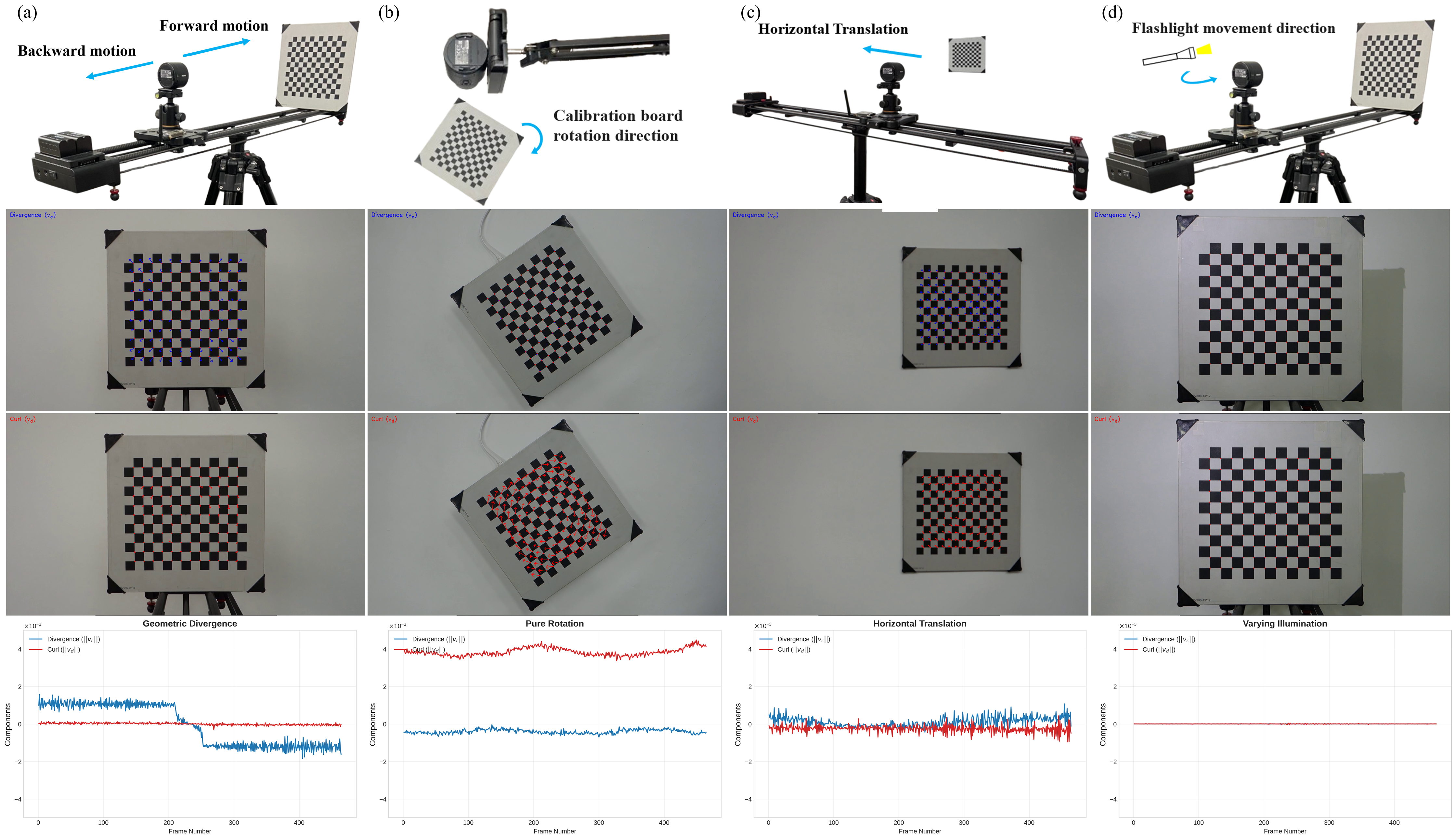}
    \caption{Top row: Experimental setups for the controlled scenarios. Second and third rows: \(\bm{v}_c\) and \(\bm{v}_d\) velocity fields with blue and red arrows. Bottom row: Temporal mean magnitudes of the CF and DF components for each scenario.}
    \label{fig:qualitative_decomposed}
\end{figure}

A Sony DSC-QX100 camera is mounted on a motorized linear slider (GVM 1.5D-120) supported by a tripod, with a \(13 \times 12\) checkerboard calibration board containing \(18\,\mathrm{mm}\) squares serving as the reference target. Camera calibration is performed using the MATLAB R2025b Camera Calibration Toolbox. Linear slider is performed at a constant speed of \(2.81\,\mathrm{cm/s}\), while the checkerboard is mounted on a motorized turntable rotating clockwise at a constant angular velocity of \(0.197\,\mathrm{rad/s}\). Camera calibration and optical alignment are summarized in~\cref{tab:setup_verification}, which reports the calibrated intrinsic parameters, lens distortion coefficients, reprojection error, and center-alignment statistics. Optical alignment is quantified by the horizontal and vertical offsets \((dx,dy)\) between the checkerboard center and the optical axis, together with the radial deviation \(r=\sqrt{dx^2+dy^2}\), reported as the mean \(\pm\) standard deviation over ten repeated trials.
Four controlled scenarios are evaluated, as shown in~\cref{fig:qualitative_decomposed}. In (a), \textit{Geometric Divergence}, the camera moves along its optical axis toward or away from the checkerboard, producing zoom-in and zoom-out effects over a camera-checkerboard distance ranging from \(58.7\) to \(82.8\,\mathrm{cm}\). In (b), \textit{Pure Rotation}, the camera remains fixed while the checkerboard rotates at a camera-checkerboard distance of \(72.3\,\mathrm{cm}\). In (c), \textit{Horizontal Translation}, the camera moves leftward along the slide rail while the checkerboard remains stationary, with the camera-checkerboard distance varying from \(88.7\) to \(93.2\,\mathrm{cm}\) at a vertical separation of \(85.5\,\mathrm{cm}\). In (d), \textit{Varying Illumination}, the camera-checkerboard distance remains fixed at \(57.4\,\mathrm{cm}\), while a light source follows a semicircular trajectory above the camera with a light-checkerboard distance ranging from \(59.0\) to \(69.3\,\mathrm{cm}\).

\begin{table}[tb]
    \centering
    \caption{Verification of the experimental setup. The left partition details the calibrated camera intrinsics, while the right partition reports the center alignment residual offsets.}
    \label{tab:setup_verification}
    \begin{adjustbox}{max width=\linewidth}
    % Increase vertical spacing (row height) by 20%
    \renewcommand{\arraystretch}{1.2}
    % Increase horizontal spacing between columns
    \setlength{\tabcolsep}{8pt}
    \begin{tabular}{@{} lcccc || lcccc @{}}
        \toprule
        \multicolumn{5}{c||}{Camera Intrinsic Parameters} & \multicolumn{5}{c}{Center Alignment Verification} \\
        \midrule
        Resolution & $f_x$ (px) & $f_y$ (px) & $c_x$ (px) & $c_y$ (px) & Case & $dx$ (px) & $dy$ (px) & $r$ (px) & $r_{\max}$ (px) \\
        $1920 \times 1080$ & 2667.4945 & 2666.5458 & 965.9914 & 537.9983 & Geometric Div. & $0.0 \pm 0.7$ & $-0.6 \pm 0.5$ & $1.0 \pm 0.4$ & $2.3$ \\
        % \addlinespace % Re-enabled to separate the intrinsic groups visually
        Repr. Error & $k_1$ & $k_2$ & $p_1$ & $p_2$ & Pure Rotation & $0.9 \pm 0.4$ & $-7.8 \pm 0.5$ & $7.9 \pm 0.5$ & $8.9$ \\
        0.0863 (px) & 0.0530 & -0.7239 & $-1.85{\times}10^{-4}$ & $4.90{\times}10^{-4}$ & Varying Illum. & $1.3 \pm 0.1$ & $-5.2 \pm 0.1$ & $5.3 \pm 0.1$ & $5.6$ \\
        \bottomrule
    \end{tabular}
    \end{adjustbox}
\end{table}

% ==== Results ==== %
\subsubsection{Results}
\label{sssec:evaluation_results}
Let $\bm{v}_i$ and $\hat{\bm{v}}_i$ denote the ground-truth and estimated flow vectors at checkerboard corner $i \in \{1,\dots,N\}$, and $\bm{v}_t$ and $\hat{\bm{v}}_t$ represent the vectors across valid frames $t \in \{1,\dots,T\}$. Spatial accuracy is measured by the standard Average Endpoint Error (AEPE) and Average Angular Error (AAE)~\cite{baker2011database}, while temporal expansion consistency and directional alignment are captured by Sparse Radial Correlation ($R_{sp}$) and Angular Cosine Similarity (ACS). Let $m_t$ and $\hat{m}_t$ denote the ground-truth and estimated radial magnitudes, with temporal means $\bar{m}$ and $\bar{\hat{m}}$, respectively. These four metrics are formulated as:
\begin{gather*}
    R_{sp} = \frac{\sum_{t=1}^T (\hat{m}_t - \bar{\hat{m}})(m_t - \bar{m})}{\sqrt{\sum_{t=1}^T (\hat{m}_t - \bar{\hat{m}})^2 \sum_{t=1}^T (m_t - \bar{m})^2}}, \quad\quad
    \mathrm{ACS} = \frac{1}{T} \sum_{t=1}^T \frac{\hat{\bm{v}}_t \cdot \bm{v}_t}{\|\hat{\bm{v}}_t\|_2 \|\bm{v}_t\|_2},\quad \\[2.5ex]
    \mathrm{AEPE} = \frac{1}{T} \sum_{t=1}^T \|\hat{\bm{v}}_t - \bm{v}_t\|_2, \:\:
    \mathrm{AAE} = \frac{1}{T} \sum_{t=1}^T \arccos \left( \frac{\hat{\bm{v}}_t \cdot \bm{v}_t + 1}{\sqrt{(\|\hat{\bm{v}}_t\|_2^2 + 1)(\|\bm{v}_t\|_2^2 + 1)}} \right).
\end{gather*}

To validate the theoretical zero-divergence and zero-curl constraints, we report repeated controlled experiments in~\cref{tab:controlled_tests}. As shown in the right panel of~\cref{fig:qualitative_decomposed}, \textit{Geometric Divergence} captures expansion-to-contraction transitions with near-zero curl, while \textit{Pure Rotation} exhibits strong curl with minimal divergence. \textit{Varying Illumination} and \textit{Horizontal Translation} produce near-zero values for both components, demonstrating robustness to rigid motion and photometric variations. Notably, the \textit{Horizontal Translation} sequence also includes slight divergence-related variation caused by perspective scaling as the checkerboard moves toward and away from the camera.

\begin{table}[tb]
    % \small
    \caption{Quantitative comparison of OF and ReynoldsFlow velocity fields across four controlled tests. Results are reported as mean $\pm$ standard deviation over 10 runs. Best results are in \textbf{bold}, second best are \underline{underlined}.}
    \label{tab:controlled_tests}
    \begin{center}
    \begin{adjustbox}{max width=\linewidth}
    \begin{tabular}{l cccc cccc}
        \toprule
        \multirow{2}{*}{Algorithms} & \multicolumn{4}{c}{Geometric Divergence} & \multicolumn{4}{c}{Varying Illumination} \\
        \cmidrule(lr){2-5} \cmidrule(lr){6-9}
        & $R_{sp} \uparrow$ & ACS $\uparrow$ & AEPE $\downarrow$ & AAE ($^\circ$) $\downarrow$ & $R_{sp} \uparrow$ & ACS $\uparrow$ & AEPE $\downarrow$ & AAE ($^\circ$) $\downarrow$ \\
        \midrule
        Horn-Schunck~\cite{horn1981determining} & \underline{0.9491}$\pm$0.08 & 0.9695$\pm$0.01 & 0.1885$\pm$0.04 & 10.3510$\pm$1.02\:\: & \underline{0.6251}$\pm$0.12 & \textbf{0.9999}$\pm$0.00 & 0.0026$\pm$0.00 & 0.1385$\pm$0.05 \\
        Lucas-Kanade~\cite{lucas1981iterative} & 0.8812$\pm$0.18 & 0.9895$\pm$0.01 & 0.1052$\pm$0.02 & 5.4210$\pm$0.48 & 0.1245$\pm$0.02 & 0.9988$\pm$0.00 & 0.0105$\pm$0.00 & 0.4512$\pm$0.06 \\
        Farneback~\cite{farneback2003two} & 0.9452$\pm$0.11 & \underline{0.9965}$\pm$0.00 & \underline{0.0635}$\pm$0.01 & \underline{3.3412}$\pm$0.29 & 0.6110$\pm$0.15 & \textbf{0.9999}$\pm$0.00 & \underline{0.0022}$\pm$0.00 & \underline{0.1221}$\pm$0.04 \\
        Brox~\cite{brox2004high} & 0.9421$\pm$0.14 & 0.9938$\pm$0.01 & 0.0832$\pm$0.02 & 4.3051$\pm$0.28 & 0.3698$\pm$0.15 & \textbf{0.9999}$\pm$0.00 & 0.0041$\pm$0.00 & 0.2125$\pm$0.05 \\
        TV-L1~\cite{zach2007duality} & 0.8781$\pm$0.19 & 0.9945$\pm$0.01 & 0.0805$\pm$0.02 & 4.1522$\pm$0.42 & 0.2791$\pm$0.08 & \underline{0.9997}$\pm$0.00 & 0.0052$\pm$0.00 & 0.2810$\pm$0.07 \\
        DeepFlow~\cite{weinzaepfel2013deepflow} & 0.8410$\pm$0.21 & 0.9928$\pm$0.01 & 0.0915$\pm$0.02 & 4.7120$\pm$0.51 & -0.1685$\pm$0.06\: & 0.9994$\pm$0.00 & 0.0195$\pm$0.00 & 1.0821$\pm$0.06 \\
        PCAFlow~\cite{wulff2015efficient} & 0.8175$\pm$0.24 & 0.9721$\pm$0.01 & 0.2054$\pm$0.02 & 11.2810$\pm$0.52\:\: & -0.0521$\pm$0.02\: & 0.9905$\pm$0.01 & 0.0892$\pm$0.03 & 4.8510$\pm$1.25 \\
        DIS~\cite{kroeger2016fast} & 0.9315$\pm$0.13 & 0.9951$\pm$0.00 & 0.0751$\pm$0.02 & 3.9850$\pm$0.38 & 0.3045$\pm$0.04 & \textbf{0.9999}$\pm$0.00 & 0.0047$\pm$0.00 & 0.2525$\pm$0.06 \\
        RPKNet~\cite{morimitsu2024recurrent} & 0.7812$\pm$0.26 & 0.9855$\pm$0.01 & 0.1465$\pm$0.02 & 7.7120$\pm$0.51 & -0.1345$\pm$0.05\: & 0.9981$\pm$0.00 & 0.0502$\pm$0.01 & 2.7650$\pm$0.42 \\
        SEA-RAFT (M)~\cite{wang2024sea} & 0.7625$\pm$0.25 & 0.9825$\pm$0.01 & 0.1652$\pm$0.02 & 9.1250$\pm$0.68 & -0.0712$\pm$0.01\: & 0.9991$\pm$0.00 & 0.0258$\pm$0.01 & 1.4150$\pm$0.28 \\
        DPFlow~\cite{morimitsu2025dpflow} & 0.8715$\pm$0.17 & 0.9835$\pm$0.01 & 0.1551$\pm$0.02 & 8.2015$\pm$0.52 & -0.1575$\pm$0.08\: & 0.9978$\pm$0.00 & 0.0581$\pm$0.02 & 3.2350$\pm$0.23 \\
        \midrule
        ReynoldsFlow (Ours) & \textbf{0.9654}$\pm$0.06 & \textbf{0.9975}$\pm$0.00 & \textbf{0.0582}$\pm$0.01 & \textbf{3.1045}$\pm$0.21 & \textbf{0.6582}$\pm$0.14 & \textbf{0.9999}$\pm$0.00 & \textbf{0.0020}$\pm$0.00 & \textbf{0.1165}$\pm$0.02 \\
        \midrule
        \midrule
        \multirow{2}{*}{Algorithms} & \multicolumn{4}{c}{Horizontal Translation} & \multicolumn{4}{c}{Pure Rotation} \\
        \cmidrule(lr){2-5} \cmidrule(lr){6-9}
        & $R_{sp} \uparrow$ & ACS $\uparrow$ & AEPE $\downarrow$ & AAE ($^\circ$) $\downarrow$ & $R_{sp} \uparrow$ & ACS $\uparrow$ & AEPE $\downarrow$ & AAE ($^\circ$) $\downarrow$ \\
        \midrule
        Horn-Schunck~\cite{horn1981determining} & 0.9154$\pm$0.11 & 0.9962$\pm$0.00 & 0.0241$\pm$0.01 & 1.2504$\pm$0.25 & 0.7021$\pm$0.09 & 0.7305$\pm$0.01 & 1.0852$\pm$0.01 & 42.0120$\pm$4.15\:\: \\
        Lucas-Kanade~\cite{lucas1981iterative} & 0.6205$\pm$0.18 & 0.9985$\pm$0.00 & 0.0165$\pm$0.00 & 0.8621$\pm$0.18 & 0.4589$\pm$0.14 & 0.9964$\pm$0.00 & 0.1235$\pm$0.01 & 3.7145$\pm$0.08 \\
        Farneback~\cite{farneback2003two} & \textbf{0.9485}$\pm$0.06 & \textbf{0.9999}$\pm$0.00 & \underline{0.0075}$\pm$0.00 & \underline{0.3910}$\pm$0.08 & 0.6924$\pm$0.10 & 0.9971$\pm$0.00 & 0.1082$\pm$0.01 & 3.4250$\pm$0.06 \\
        Brox~\cite{brox2004high} & 0.8950$\pm$0.15 & 0.9995$\pm$0.00 & 0.0112$\pm$0.00 & 0.5621$\pm$0.12 & \underline{0.8582}$\pm$0.08 & 0.9975$\pm$0.00 & 0.1271$\pm$0.01 & 3.4102$\pm$0.07 \\
        TV-L1~\cite{zach2007duality} & 0.8675$\pm$0.12 & 0.9992$\pm$0.00 & 0.0102$\pm$0.00 & 0.5422$\pm$0.14 & 0.5235$\pm$0.13 & 0.9968$\pm$0.00 & 0.1050$\pm$0.01 & 3.3055$\pm$0.11 \\
        DeepFlow~\cite{weinzaepfel2013deepflow} & 0.3905$\pm$0.21 & 0.9989$\pm$0.00 & 0.0275$\pm$0.00 & 1.4510$\pm$0.31 & 0.6558$\pm$0.13 & \underline{0.9981}$\pm$0.00 & \underline{0.0845}$\pm$0.01 & \underline{2.6512}$\pm$0.08 \\
        PCAFlow~\cite{wulff2015efficient} & 0.5125$\pm$0.23 & 0.9972$\pm$0.00 & 0.0381$\pm$0.01 & 2.0215$\pm$0.48 & 0.1685$\pm$0.08 & 0.9925$\pm$0.01 & 0.1945$\pm$0.01 & 5.8420$\pm$0.12 \\
        DIS~\cite{kroeger2016fast} & 0.8621$\pm$0.16 & 0.9991$\pm$0.00 & 0.0115$\pm$0.00 & 0.5512$\pm$0.11 & 0.4610$\pm$0.09 & 0.9969$\pm$0.00 & 0.1095$\pm$0.01 & 3.4510$\pm$0.07 \\
        RPKNet~\cite{morimitsu2024recurrent} & 0.3485$\pm$0.25 & 0.9974$\pm$0.00 & 0.0441$\pm$0.01 & 2.3510$\pm$0.52 & 0.0791$\pm$0.11 & 0.9912$\pm$0.01 & 0.2485$\pm$0.01 & 7.0325$\pm$0.11 \\
        SEA-RAFT (M)~\cite{wang2024sea} & 0.2745$\pm$0.28 & 0.9971$\pm$0.00 & 0.0565$\pm$0.01 & 3.0750$\pm$0.68 & 0.0415$\pm$0.09 & 0.9915$\pm$0.01 & 0.1925$\pm$0.02 & 6.1050$\pm$0.52 \\
        DPFlow~\cite{morimitsu2025dpflow} & 0.4855$\pm$0.24 & 0.9976$\pm$0.00 & 0.0472$\pm$0.01 & 2.5910$\pm$0.65 & 0.1145$\pm$0.12 & 0.9942$\pm$0.01 & 0.1785$\pm$0.01 & 5.5510$\pm$0.10 \\
        \midrule
        ReynoldsFlow (Ours) & \underline{0.9402}$\pm$0.08 & \underline{0.9998}$\pm$0.00 & \textbf{0.0064}$\pm$0.00 & \textbf{0.3651}$\pm$0.05 & \textbf{0.9015}$\pm$0.03 & \textbf{0.9993}$\pm$0.00 & \textbf{0.0615}$\pm$0.01 & \textbf{1.9840}$\pm$0.07 \\
        \bottomrule
    \end{tabular}
    \end{adjustbox}
    \end{center}
\end{table}

% ======== Pose Estimation on GolfDB ======== %
\subsection{Pose Estimation on GolfDB}
\label{ssec:golfdb}
We evaluate pose estimation on GolfDB~\cite{9025559}, which contains 1,400 videos at \(160\times160\) resolution with subjects occupying nearly the full frame. Each golf swing is annotated with eight events: Address (A), Toe-up (TU), Mid-backswing (MB), Top (T), Mid-downswing (MD), Impact (I), Mid-follow-through (MFT), and Finish (F), captured from face-on and down-the-line views.
We use SwingNet~\cite{9025559} with a setup consisting of sequence length 64, 10 frozen layers, batch size 22, 2,000 iterations, six workers, initializing with the pre-trained MobileNetV2~\cite{sandler2018mobilenetv2} and fine-tuning on GolfDB. Performance is measured using the Percentage of Correct Events (PCE) with four-fold cross-validation. For real-time 30\,fps videos, we use a tolerance of $\delta = 1$; for slow-motion, $\delta = \max(\lfloor \frac{N}{f} \rceil, 1)$, where $N$ is the number of frames from Address to Impact and $f$ the sampling frequency. As shown in~\cref{tab:results}, our input achieves the highest PCE of 0.812.

\begin{table}[tb]
    % \small
    \caption{Effect of velocity fields estimation on downstream task performance across GolfDB~\cite{9025559}, HMDB51~\cite{kuehne2011hmdb}, UCF101~\cite{soomro2012ucf101}, Anti-UAV~\cite{jiang2021anti}, ARD100~\cite{guo2025yolomg}, and UAVDB~\cite{chen2024uavdb}. Best results are in \textbf{bold}, second best are \underline{underlined}.}
    \label{tab:results}
    \begin{center}
    \begin{adjustbox}{max width=\linewidth, max height=0.5\textheight}
    \begin{tabular}{lccccccccc}
        \toprule
        \multirow{2.5}{*}{Methods} & GolfDB & HMDB51 & UCF101 & \multicolumn{2}{c}{Anti-UAV} & \multicolumn{2}{c}{ARD100} & \multicolumn{2}{c}{UAVDB} \\
        \cmidrule(lr){2-2} \cmidrule(lr){3-3} \cmidrule(lr){4-4} \cmidrule(lr){5-6} \cmidrule(lr){7-8} \cmidrule(lr){9-10}
        & PCE \(\uparrow\) & Accuracy \(\uparrow\) & Accuracy \(\uparrow\) & \(\text{AP}^{test}_{50}\) \(\uparrow\) & \(\text{AP}^{test}_{50-95}\) \(\uparrow\) & \(\text{AP}^{test}_{50}\) \(\uparrow\) & \(\text{AP}^{test}_{50-95}\) \(\uparrow\) & \(\text{AP}^{test}_{50}\) \(\uparrow\) & \(\text{AP}^{test}_{50-95}\) \(\uparrow\) \\
        \midrule
        RGB & 0.705 & 0.372 & 0.698 & \multicolumn{2}{c}{--} & \underline{0.554} & \underline{0.304} & \underline{0.811} & \underline{0.518}\\
        Grayscale / Infrared & 0.698 & 0.328 & 0.614 & \underline{0.781} & \underline{0.418} & 0.376 & 0.167 & 0.660 & 0.281\\
        Horn-Schunck~\cite{horn1981determining} & 0.707 & 0.165 & 0.259 & 0.217 & 0.080 & 0.074 & 0.016 & 0.104 & 0.021\\
        Lucas-Kanade~\cite{lucas1981iterative} & 0.692 & 0.214 & 0.338 & 0.280 & 0.114 & 0.237 & 0.141 & 0.500 & 0.200\\
        Farneback~\cite{farneback2003two} & 0.717 & 0.248 & 0.383 & 0.500 & 0.246 & 0.182 & 0.103 & 0.258 & 0.145\\
        Brox~\cite{brox2004high} & 0.708 & 0.260 & 0.328 & 0.379 & 0.168 & 0.122 & 0.089 & 0.244 & 0.110\\
        TV-L1~\cite{zach2007duality} & \underline{0.810} & 0.284 & 0.537 & 0.600 & 0.278 & 0.227 & 0.127 & 0.779 & 0.409\\
        DeepFlow~\cite{weinzaepfel2013deepflow} & 0.706 & 0.237 & 0.419 & 0.344 & 0.143 & 0.138 & 0.063 & 0.154 & 0.058\\
        PCAFlow~\cite{wulff2015efficient} & 0.765 & 0.296 & 0.424 & 0.580 & 0.286 & 0.128 &0.041 & 0.547 & 0.332 \\
        DIS~\cite{kroeger2016fast} & 0.772 & 0.207 & 0.562 & 0.347 & 0.144 & 0.102 & 0.018 & 0.151 & 0.057\\
        RPKNet~\cite{morimitsu2024recurrent} & 0.801 & 0.395 & \textbf{0.734} & 0.316 & 0.214 & 0.088 & 0.014 & 0.110 & 0.039\\
        SEA-RAFT (M)~\cite{wang2024sea} & 0.782 & \textbf{0.419} & \underline{0.722} & 0.357 & 0.188 & 0.089 & 0.048 & 0.486 & 0.243\\
        DPFlow~\cite{morimitsu2025dpflow} & 0.786 & \underline{0.411} & 0.705 & 0.427 & 0.265 & 0.072 & 0.016 & 0.270 & 0.101\\
        \midrule
        ReynoldsFlow (Ours) & \textbf{0.812} & 0.402 & 0.714 & \textbf{0.792} & \textbf{0.446} & \textbf{0.602} & \textbf{0.326} & \textbf{0.895} & \textbf{0.547}\\
        \bottomrule
    \end{tabular}
    \end{adjustbox}
    \end{center}
\end{table}

% ======== Robustness to Geometric and Illumination Variations ======== %
\subsection{Action Recognition on HMDB51 and UCF101}
\label{ssec:action_recognition}
Beyond pose estimation, we evaluate action recognition on two widely used benchmarks: HMDB51~\cite{kuehne2011hmdb}, containing 6,766 clips across 51 action categories (resolutions 176\(\times\)240–592\(\times\)240), and UCF101~\cite{soomro2012ucf101}, with 13,320 clips across 101 categories at 320\(\times\)240. 
Following the self-supervised video representation framework of~\cite{jenni2020video}, we adopt their C3D-based architecture with different flow inputs, keeping the original training configuration (batch size 6, 100 epochs pre-training, 75 epochs fine-tuning). Classification accuracy is reported using two-fold cross-validation. As shown in~\cref{tab:results}, ReynoldsFlow consistently improves over the RGB baseline and performs comparably to DL-based methods, even if it does not achieve the highest absolute accuracy.

% ======== Object Detection on UAV Datasets ======== %
\subsection{Object Detection on UAV Datasets}
\label{ssec:uavdb}
We evaluate ReynoldsFlow on UAV object detection using three datasets: Anti-UAV, ARD100, and UAVDB. Anti-UAV contains infrared frames (512\(\times\)512 to 640\(\times\)512) captured by a moving camera; ARD100 comprises high-resolution RGB videos (1920\(\times\)1080) from Phantom-type drones; UAVDB includes high-resolution RGB frames (1920\(\times\)1080 to 3840\(\times\)2160) from a static ground camera. All datasets focus on single-class UAVs at varying scales. We sample 4,800/ 1,600/1,600 images for training/validation/test from Anti-UAV, approximately 12,000/4,000/4,000 from ARD100, and use the official split for UAVDB.

Flow estimates from all methods are fed into YOLOv11n~\cite{yolo11_ultralytics}, trained with a batch size of 64, 640\(\times\)640 input resolution, eight workers, and over 100 epochs. Mosaic augmentation is applied except for the final ten epochs, and pre-trained weights are fine-tuned for each dataset. Performance is reported using $\text{AP}_{50}$ and $\text{AP}_{50-95}$. As shown in~\cref{tab:results}, ReynoldsFlow consistently improves YOLOv11n across all datasets.
Beyond~\cref{tab:results}, ReynoldsFlow achieves comparable or higher accuracy than spatiotemporal embedding models, such as TransVisDrone~\cite{sangam2022transvisdrone}, which are specifically designed for UAV detection. However, these models require over five times longer training and more than six times slower inference. In contrast, ReynoldsFlow adds only negligible overhead relative to model inference, avoids task-specific retraining, and seamlessly integrates as a plug-and-play representation for diverse downstream tasks.

% ======== Analysis and Discussion ======== %
\subsection{Analysis and Discussion}
\label{ssec:analysis_and_discussion}
ReynoldsFlow demonstrates clear advantages in accuracy, efficiency, and representation design. Large objects benefit from directional cues, yielding predictable gains across flow features. The strength of ReynoldsFlow is especially pronounced for tiny or nearly invisible targets, where directional information is less informative but motion magnitude remains critical, allowing ReynoldsFlow to maintain high accuracy where both classical and DL methods struggle.

\begin{table}[tb]
    % \small
    \caption{Ablation study of ReynoldsFlow representations on downstream task performance. Best results are in \textbf{bold}, second best are \underline{underlined}.}
    \label{tab:ablation}
    \begin{center}
    \begin{adjustbox}{max width=\linewidth, max height=0.5\textheight}
    \begin{tabular}{lccccccccc}
        \toprule
        \multirow{2.5}{*}{Representations} & GolfDB & HMDB51 & UCF101 & \multicolumn{2}{c}{Anti-UAV} & \multicolumn{2}{c}{ARD100} & \multicolumn{2}{c}{UAVDB} \\
        \cmidrule(lr){2-2} \cmidrule(lr){3-3} \cmidrule(lr){4-4} \cmidrule(lr){5-6} \cmidrule(lr){7-8} \cmidrule(lr){9-10}
        & PCE \(\uparrow\) & Accuracy \(\uparrow\) & Accuracy \(\uparrow\) & \(\text{AP}^{test}_{50}\) \(\uparrow\) & \(\text{AP}^{test}_{50-95}\) \(\uparrow\) & \(\text{AP}^{test}_{50}\) \(\uparrow\) & \(\text{AP}^{test}_{50-95}\) \(\uparrow\) & \(\text{AP}^{test}_{50}\) \(\uparrow\) & \(\text{AP}^{test}_{50-95}\) \(\uparrow\) \\
        \midrule
        HSV~\cite{liu2009beyond} & \underline{0.804} & \underline{0.382} & 0.597 & 0.646 & 0.320 & 0.417 & 0.211 & 0.500 & 0.288 \\
        \(\left[|\bm{v}_d^n|, \:\:-\:\:, f^n\right]\) & 0.788 & 0.375 & 0.684 & 0.765 & \underline{0.407} & 0.478 & 0.262 & 0.803 & 0.471 \\
        \(\left[|\bm{v}_c^n|, |\bm{v}_d^n|, f^n\right]\) & 0.791 & 0.367 & \underline{0.699} & \underline{0.784} & 0.386 & \underline{0.509} & \underline{0.287} & \underline{0.831} & \underline{0.492} \\
        \midrule
        \(\left[|\bm{v}_d^n|, |\bm{v}_c^n|, f^n\right]\) & \textbf{0.812} & \textbf{0.402} & \textbf{0.714} & \textbf{0.792} & \textbf{0.446} & \textbf{0.602} & \textbf{0.326} & \textbf{0.895} & \textbf{0.547}\\
        \bottomrule
    \end{tabular}
    \end{adjustbox}
    \end{center}
\end{table}

Our ablation study provides three key insights. First, the proposed ReynoldsFlow representation consistently outperforms conventional HSV-based visualizations by omitting directional encoding, as shown in the first and last rows of~\cref{tab:ablation}. Second, incorporating the CF magnitude \(|\bm{v}_c^n|\) consistently improves performance, as demonstrated by the second and last rows. Third, among six channel-ordering configurations, assigning motion magnitude to the green channel achieves the best performance, consistent with the RGGB Bayer pattern, in which the green channel has the highest sampling density and contributes most to luminance perception. These findings highlight the importance of the CF component and the proposed feature representation.

% ================ Conclusion ================ %
\section{Conclusion}
\label{sec:conclusion}
In this paper, we presented ReynoldsFlow, a physics-inspired, zero-shot video representation that overcomes the brightness constancy limitation in motion estimation. By grounding our formulation in the RTT and HHD, we decompose video motion into CF and DF components, yielding a divergence-compensated solver that eliminates tracking artifacts caused by geometric zooming and varying illumination. 
Quantitative results show that ReynoldsFlow consistently outperforms classical and DL-based OF methods in challenging scenarios while running in real time. As a robust, plug-and-play representation, it stacks the CF and DF magnitudes to enhance downstream tasks, including pose estimation, action recognition, and tiny UAV detection, without task-specific retraining or costly 3D convolutions. 
ReynoldsFlow demonstrates that transparent, physics-grounded modeling can provide an interpretable, efficient, and powerful alternative to purely data-driven architectures for video understanding.

% \clearpage  % TODO FINAL: This \clearpage needs to be removed from both review and camera-ready versions.

% ================ Acknowledgements ================ %
\section*{Acknowledgements}
This work was supported in part by the resources provided by National Yang Ming Chiao Tung University and The University of Melbourne. The authors also acknowledge the developers of the open-source software and the providers of the publicly available datasets that made this research possible. The corresponding author gratefully acknowledges the support from the National Science and Technology Council (NSTC), Taiwan, under Grant No.~114-2115-M-008-007-MY2.

\bibliographystyle{splncs04}
\bibliography{main}
\end{document}